\documentclass{article}
\usepackage{arxiv}
\usepackage{times}
\usepackage{epsfig}
\usepackage{graphicx}
\usepackage{amsmath}
\usepackage{amssymb}
\usepackage{tikz}
\usepackage{ifthen}
\usepackage{scalefnt}
\usepackage{placeins}
\usepackage{microtype}
\usepackage{graphicx}
\usepackage{subfigure}
\usepackage{booktabs}
\usepackage{algorithm,algpseudocode}
\usepackage{amsmath}
\usepackage{amsfonts}
\usepackage{amssymb}
\usepackage{fixltx2e}
\usepackage{multirow}
\usepackage{amsmath}
\usepackage{dblfloatfix}
\usepackage{multicol}
\usepackage{lipsum}

\newcommand\blfootnote[1]{%
  \begingroup
  \renewcommand\thefootnote{}\footnote{#1}%
  \addtocounter{footnote}{-1}%
  \endgroup
}
\usepackage{hyperref}
\usepackage{mathrsfs}
\usepackage{enumitem}

\usepackage{scalerel,stackengine}
\stackMath
\newcommand\reallywidehat[1]{%
\savestack{\tmpbox}{\stretchto{%
  \scaleto{%
    \scalerel*[\widthof{\ensuremath{#1}}]{\kern-.6pt\bigwedge\kern-.6pt}%
    {\rule[-\textheight/2]{1ex}{\textheight}}
  }{\textheight}%
}{0.5ex}}%
\stackon[1pt]{#1}{\tmpbox}%
}

\usepackage{algorithmicx}
\usepackage{algorithm}
\usepackage{algpseudocode}

\usetikzlibrary{decorations.pathreplacing}

\newif\ifcuboidshade
\newif\ifcuboidemphedge

\tikzset{
	cuboid/.is family,
	cuboid,
	shiftx/.initial=0,
	shifty/.initial=0,
	dimx/.initial=3,
	dimy/.initial=3,
	dimz/.initial=3,
	scale/.initial=1,
	densityx/.initial=1,
	densityy/.initial=1,
	densityz/.initial=1,
	rotation/.initial=0,
	anglex/.initial=0,
	angley/.initial=90,
	anglez/.initial=225,
	scalex/.initial=1,
	scaley/.initial=1,
	scalez/.initial=0.5,
	front/.style={draw=black,fill=white},
	top/.style={draw=black,fill=white},
	right/.style={draw=black,fill=white},
	shade/.is if=cuboidshade,
	shadecolordark/.initial=black,
	shadecolorlight/.initial=white,
	shadeopacity/.initial=0.15,
	shadesamples/.initial=16,
	emphedge/.is if=cuboidemphedge,
	emphstyle/.style={thick},
}

\newcommand{\tikzcuboidkey}[1]{\pgfkeysvalueof{/tikz/cuboid/#1}}

\newcommand{\tikzcuboid}[1]{
	\tikzset{cuboid,#1} 
	\pgfmathsetlengthmacro{\vectorxx}{\tikzcuboidkey{scalex}*cos(\tikzcuboidkey{anglex})*28.452756}
	\pgfmathsetlengthmacro{\vectorxy}{\tikzcuboidkey{scalex}*sin(\tikzcuboidkey{anglex})*28.452756}
	\pgfmathsetlengthmacro{\vectoryx}{\tikzcuboidkey{scaley}*cos(\tikzcuboidkey{angley})*28.452756}
	\pgfmathsetlengthmacro{\vectoryy}{\tikzcuboidkey{scaley}*sin(\tikzcuboidkey{angley})*28.452756}
	\pgfmathsetlengthmacro{\vectorzx}{\tikzcuboidkey{scalez}*cos(\tikzcuboidkey{anglez})*28.452756}
	\pgfmathsetlengthmacro{\vectorzy}{\tikzcuboidkey{scalez}*sin(\tikzcuboidkey{anglez})*28.452756}
	\begin{scope}[xshift=\tikzcuboidkey{shiftx}, yshift=\tikzcuboidkey{shifty}, scale=\tikzcuboidkey{scale}, rotate=\tikzcuboidkey{rotation}, x={(\vectorxx,\vectorxy)}, y={(\vectoryx,\vectoryy)}, z={(\vectorzx,\vectorzy)}]
		\pgfmathsetmacro{\steppingx}{1/\tikzcuboidkey{densityx}}
		\pgfmathsetmacro{\steppingy}{1/\tikzcuboidkey{densityy}}
		\pgfmathsetmacro{\steppingz}{1/\tikzcuboidkey{densityz}}
		\newcommand{\dimx}{\tikzcuboidkey{dimx}}
		\newcommand{\dimy}{\tikzcuboidkey{dimy}}
		\newcommand{\dimz}{\tikzcuboidkey{dimz}}
		\pgfmathsetmacro{\secondx}{2*\steppingx}
		\pgfmathsetmacro{\secondy}{2*\steppingy}
		\pgfmathsetmacro{\secondz}{2*\steppingz}
		\ifthenelse{\equal{\dimx}{1}}
		{\foreach \x in {\steppingx,...,\dimx}}
		{\foreach \x in {\steppingx,\secondx,...,\dimx}}
		{     \ifthenelse{\equal{\dimy}{1}}
			{\foreach \y in {\steppingy,...,\dimy}}
			{\foreach \y in {\steppingy,\secondy,...,\dimy}}
			{   \pgfmathsetmacro{\lowx}{(\x-\steppingx)}
				\pgfmathsetmacro{\lowy}{(\y-\steppingy)}
				\filldraw[cuboid/front] (\lowx,\lowy,\dimz) -- (\lowx,\y,\dimz) -- (\x,\y,\dimz) -- (\x,\lowy,\dimz) -- cycle;
			}
		}
		\ifthenelse{\equal{\dimx}{1}}
		{\foreach \x in {\steppingx,...,\dimx}}
		{\foreach \x in {\steppingx,\secondx,...,\dimx}}
		{ \ifthenelse{\equal{\dimz}{1}}
			{\foreach \z in {\steppingz,...,\dimz}}
			{\foreach \z in {\steppingz,\secondz,...,\dimz}}
			{   \pgfmathsetmacro{\lowx}{(\x-\steppingx)}
				\pgfmathsetmacro{\lowz}{(\z-\steppingz)}
				\filldraw[cuboid/top] (\lowx,\dimy,\lowz) -- (\lowx,\dimy,\z) -- (\x,\dimy,\z) -- (\x,\dimy,\lowz) -- cycle;
			}
		}
		\ifthenelse{\equal{\dimy}{1}}
		{\foreach \y in {\steppingy,...,\dimy}}
		{\foreach \y in {\steppingy,\secondy,...,\dimy}}
		{ \ifthenelse{\equal{\dimz}{1}}
			{\foreach \z in {\steppingz,...,\dimz}}
			{\foreach \z in {\steppingz,\secondz,...,\dimz}}
			{   \pgfmathsetmacro{\lowy}{(\y-\steppingy)}
				\pgfmathsetmacro{\lowz}{(\z-\steppingz)}
				\filldraw[cuboid/right] (\dimx,\lowy,\lowz) -- (\dimx,\lowy,\z) -- (\dimx,\y,\z) -- (\dimx,\y,\lowz) -- cycle;
			}
		}
		\ifcuboidemphedge
		\draw[cuboid/emphstyle] (0,\dimy,0) -- (\dimx,\dimy,0) -- (\dimx,\dimy,\dimz) -- (0,\dimy,\dimz) -- cycle;%
		\draw[cuboid/emphstyle] (0,\dimy,\dimz) -- (0,0,\dimz) -- (\dimx,0,\dimz) -- (\dimx,\dimy,\dimz);%
		\draw[cuboid/emphstyle] (\dimx,\dimy,0) -- (\dimx,0,0) -- (\dimx,0,\dimz);%
		\fi
		
		\ifcuboidshade
		\pgfmathsetmacro{\cstepx}{\dimx/\tikzcuboidkey{shadesamples}}
		\pgfmathsetmacro{\cstepy}{\dimy/\tikzcuboidkey{shadesamples}}
		\pgfmathsetmacro{\cstepz}{\dimz/\tikzcuboidkey{shadesamples}}
		\foreach \s in {1,...,\tikzcuboidkey{shadesamples}}
		{   \pgfmathsetmacro{\lows}{\s-1}
			\pgfmathsetmacro{\cpercent}{(\lows)/(\tikzcuboidkey{shadesamples}-1)*100}
			\fill[opacity=\tikzcuboidkey{shadeopacity},color=\tikzcuboidkey{shadecolorlight}!\cpercent!\tikzcuboidkey{shadecolordark}] (0,\s*\cstepy,\dimz) -- (\s*\cstepx,\s*\cstepy,\dimz) -- (\s*\cstepx,0,\dimz) -- (\lows*\cstepx,0,\dimz) -- (\lows*\cstepx,\lows*\cstepy,\dimz) -- (0,\lows*\cstepy,\dimz) -- cycle;
			\fill[opacity=\tikzcuboidkey{shadeopacity},color=\tikzcuboidkey{shadecolorlight}!\cpercent!\tikzcuboidkey{shadecolordark}] (0,\dimy,\s*\cstepz) -- (\s*\cstepx,\dimy,\s*\cstepz) -- (\s*\cstepx,\dimy,0) -- (\lows*\cstepx,\dimy,0) -- (\lows*\cstepx,\dimy,\lows*\cstepz) -- (0,\dimy,\lows*\cstepz) -- cycle;
			\fill[opacity=\tikzcuboidkey{shadeopacity},color=\tikzcuboidkey{shadecolorlight}!\cpercent!\tikzcuboidkey{shadecolordark}] (\dimx,0,\s*\cstepz) -- (\dimx,\s*\cstepy,\s*\cstepz) -- (\dimx,\s*\cstepy,0) -- (\dimx,\lows*\cstepy,0) -- (\dimx,\lows*\cstepy,\lows*\cstepz) -- (\dimx,0,\lows*\cstepz) -- cycle;
		}
		\fi 
		
	\end{scope}
}

\begin{document}

\twocolumn[
\title{MUSCO: Multi-Stage Compression of neural networks }

\begin{center}
\author{
Julia Gusak\textsuperscript{1*},\quad
Maksym Kholiavchenko\textsuperscript{2*},\quad
Evgeny Ponomarev\textsuperscript{1},\quad
Larisa Markeeva\textsuperscript{1},\\
{\bf Andrzej Cichocki,\textsuperscript{1}\quad} 
{\bf Ivan Oseledets\textsuperscript{1}}\\
\\
\textsuperscript{1}{Skolkovo Institute of Science and Technology, Moscow, Russia} \and
\textsuperscript{2}{Innopolis University, Kazan, Russia}
\\
\small e-mails: \{\textit{y.gusak, evgenii.ponomarev, l.markeeva, a.cichocki, i.oseledets}\}@skoltech.ru,\\ \small \textit{m.kholyavchenko}@innopolis.ru
}
\end{center}

\maketitle
]

\begin{abstract}
The low-rank tensor approximation is very promising for the compression of deep neural networks. We propose a new simple and efficient iterative approach, which alternates low-rank factorization with a smart rank selection and fine-tuning. We demonstrate the efficiency of our method comparing to non-iterative ones. Our approach improves the compression rate while maintaining the accuracy for a variety of tasks. 
\end{abstract}

\section{Introduction}
\blfootnote{* Contributed equally.}
The development of deeper and more complex networks in order to achieve higher performance has become commonplace. However such networks contain tens of millions of parameters and often cannot be efficiently deployed on embedded systems and mobile devices due to their computational power and memory limitations.

Low-rank matrix and tensor approximations provide excellent compression of neural network layers \cite{jaderberg2014speeding, kim2015compression, lebedev2014speeding, zhang2016accelerating}. 
In these methods, factorization of weight tensors yields an approximate compressed network. 
For example, when we approximate a 4-dimensional convolutional kernel by a tensor, whose factorized form has three components,  we can replace a corresponding layer  with three consecutive convolutional layers (Figure~\ref{fig:tucker2_compress}). However, approaches based on a low-rank tensor factorization are built on the same scheme: compression followed by fine-tuning to compensate for a significant loss of the quality of the model. The main benefit of this approach is that the compressed version provides an initial approximation, which leads to a better quality after fine-tuning than if the same architecture is learned directly from a  random initialization. 

In our paper, we propose a way to substantially improve the abovementioned scheme by  applying low-rank decomposition and fine-tuning iteratively several times (Algorithm~\ref{alg:compression}). It turns out that such simple idea can significantly improve the quality of neural networks compression. For example, for Faster R-CNN  with  ResNet-50 backbone, we achieve a better compressed model for the same quality than can be  obtained with non-iterative algorithm (Section~\ref{sec:experiments}). 

\begin{figure}[tb]
\includegraphics[width=1.0\linewidth]{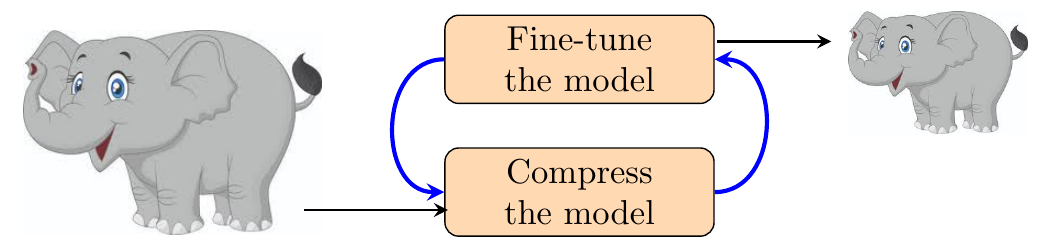}
\centering
\caption{Iterative compression. Input: pre-trained original model. Output: fine-tuned compressed model. At each iteration every layer is (further) compressed via a low-rank approximation of its weights tensor.}
\label{fig:compression_algo}
\end{figure}

We introduce \emph{Multi-Stage COmpression method (MUSCO)} for automated network compression
(Sections~\ref{sec:problem_statement},~\ref{sec:compression_algo}). The algorithm consists of two repetitive steps: compression and fine-tuning (Figure~\ref{fig:compression_algo}). 
During the compression step model weights of selected layers are approximated using tensor decomposition with automatically selected rank values (Section~\ref{sec:rank_selection}). 
At this step, the redundancy present in the weight parameters is partially reduced.
The next step allows to recover the quality of the model by performing fine-tuning. By repeating these two steps several times we can gradually compress the model by substantially reducing the number of parameters in the selected layers.  

In comparison with other approaches, MUSCO does not loose quality significantly during more agressive parameters reduction. In practice, MUSCO allows achieving higher compression ratio than state-of-the-art non-iterative approaches with the same quality of the model.
Our main contributions are:
\begin{itemize}[leftmargin=*]
\item We propose an iterative low-rank approximation algorithm to efficiently compress neural networks that outperform non-iterative methods for the desired accuracy.
\item We introduce a method for automatic tensor rank selection for tensor approximations performed at each compression step.
\item We validate and demonstrate the high efficiency of our approach in a series of extensive computational experiments for object detection and classification problems.
\end{itemize}

\section{Problem statement}\label{sec:problem_statement}
{\bf In this section we introduce a formal description of a model compression in terms of transitions from one class of models to another.}  


Each neural network can be described as a pair $(f, \theta)$, where $\theta$ denotes model parameters and $f$ defines network architecure (i.e. graph structure).  Given $\theta$, a continuous function $f$ assigns to each input $X$ a result of its propagation through the whole network, $f(X, \theta)$.  

Let $M$ be our pre-trained neural network model.
We denote the class of all neural networks with the same architecture as $M$ by
${\mathcal{M} = \{(f, \theta) | \theta \in \Theta\}}$. Here $\theta$ is an array of weight tensors that parametrize individual layers of the network  architecture $f$, and $\Theta$ defines a set of all possible parametrizations. 

We perform a network compression via low-rank tensor approximation of its weight tensors $\theta$. The concept of \textit{rank} can be defined for any tensor. We use  $\mathrm{rank}(\theta)$ to determine an array of ranks corresponding to tensors in $\theta$. The expression $\mathrm{rank}(\theta) \le R$, where $R$ is an array of constant values, is used to describe elementwise constraints applied to tensors from $\theta$. 

To apply a \textit{rank-$R$  factorization} to the weights ${\theta \in \Theta}$ is to find an array of $\hat{\theta} $ from $\Theta^R$,
\begin{equation}
    \Theta^R = \{\theta \in \Theta \mid \text{rank}(\theta) \le R\},
\end{equation}
which approximates $\theta$ in a certain norm and can be represented in a factorized format $\hat{\theta}_{\mathrm{fact}} \in \Theta^R_{\mathrm{fact}}$ (i.e. $\hat{\theta}_{\mathrm{fact}}$ is an array, where each element corresponds to one weight tensor from~$\theta$ and represented by a tuple of factors). 

We introduce operators $\mathcal{F}_{\mathrm{fact}}$ and $\mathcal{F}_{\mathrm{full}}$ that perform these mappings from $\Theta^R$ to $\Theta^R_{\mathrm{fact}}$ and vice versa, i.e. 
\begin{equation}
    \Theta^R_{\mathrm{fact}} = \{ \mathcal{F}_{\mathrm{fact}}(\theta) \mid \theta \in \Theta^R\}
\end{equation}
and $\mathcal{F}_{\mathrm{full}}(\mathcal{F}_{\mathrm{fact}}(\theta)) = \theta$ for $\theta\in\Theta^R$.

\begin{figure}[!h]
\includegraphics[width=.9\linewidth]{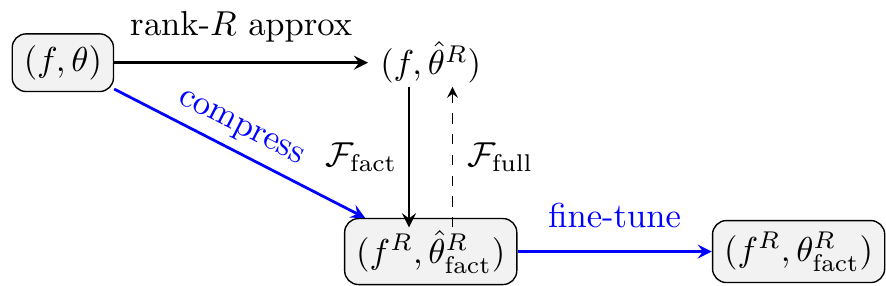}
\centering
\caption{Iterative compression (first iteration): pre-trained weights $\theta$ are factorized and corresponding linear layers of the initial architecture $f$ are decomposed into sequences of linear layers, which yield a new architecture $f^R$. (A number of layers in each sequence is equal to the number of components in the tensor factorization.)}
\label{img:compress}
\end{figure}

{\bf When we compress a network 
${(f, \theta) \in \mathcal{M}}$ using a $\mathbf{\text{rank-}R}$ weight factorization} (Figure~\ref{img:compress}),
 firstly, we obtain a model $(f, \hat{\theta})$ with the same arhitecture by projecting  $\theta$ to the parameter set $\Theta^R \subseteq \Theta$. Secondly, we  get
a compressed model $(f^R, \hat{\theta}_{\mathrm{fact}}) \in \mathcal{M}_R$ with a new architecture $f^R$  by replacing $\hat{\theta}$ with its factorized version $\hat{\theta}_{\mathrm{fact}}$, where \begin{equation}
\mathcal{M}_R = \{(f^R, \theta) \mid \theta \in \Theta^R_{\mathrm{fact}}\}    
\end{equation}
and $f^R$ is a modification of  $f$,  which contains decomposed linear layers instead of original ones. A \textit{decomposed layer} is a sequence of linear layers, each of which is represented by one factor (component) from the factorization of the original layer weight tensor. 

{\bf After fine-tuning a network 
$(f^R, \hat{\theta}_{\mathrm{fact}})$}, 
we obtain a model ${\left(f^R, \theta_{\mathrm{fact}}\right)\in \mathcal{M}_R}$, which attains a local minimum of the loss function calculated on  training samples. 

\begin{figure}[!h]
\includegraphics[width=1.\linewidth]{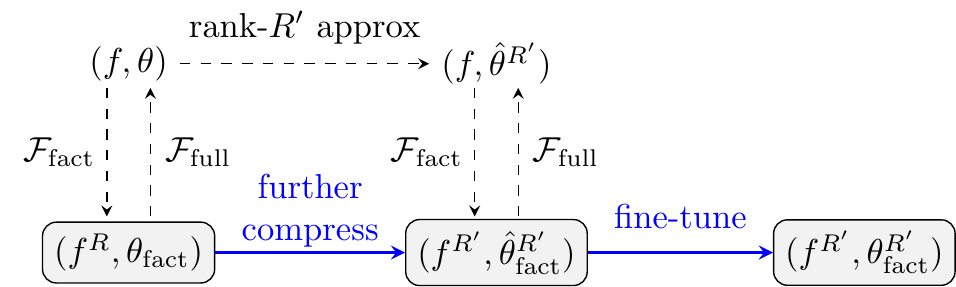}
\centering
\caption{Iterative compression (next iteration): the rank of factorized weights $\theta_{\mathrm{fact}}$ is further reduced. The architecture is changed but the number of linear layers remains the same.}
\label{img:compress_further}
\end{figure}

{\bf When we further compress an already decomposed network ${\left(f^R, \theta_{\mathrm{fact}}\right)}$},  we apply the mentioned rank-$R'$  compression procedure   to  the model ${\left(f, \theta\right) \in \mathcal{M}}$, where ${\theta \in \Theta^R}$ (Figure~\ref{img:compress_further}). In Section~\ref{sec:compress_one} we show how this step can be optimized for different types of tensor factorizations.

{\bf Thus, when we gradully compress the pre-trained model $\mathbf{M}$  over $\mathbf{K}$ iterations} (each iteration contains compression and fine-tuning steps),  we sequentially obtain  models from classes $\mathcal{M}_{R_1},\dots.\mathcal{M}_{R_K}$, namely,
\begin{equation} \label{eq:model_sequence}
    \begin{gathered}
    M \xrightarrow{} 
    \reallywidehat{M}_1 
    \xrightarrow{}
    M_1\xrightarrow{} \dots \xrightarrow{}\reallywidehat{M}_K \xrightarrow{} M_K,\\
    \text{s.t.}\quad \reallywidehat{M}_k, M_k \in \mathcal{M}_{R_k},\\
    \end{gathered}
\end{equation}
where $R_1 \ge ...\ge R_K$.
Transitions $M_{k-1} \xrightarrow{} \reallywidehat{M}_k$ and $\reallywidehat{M}_k \xrightarrow{}M_k$ correspond to the compression and fine-tuning steps respectively, $k = 1...K$. 
To compare, for non-iterative approach the similar path  looks like $M \xrightarrow{}\reallywidehat{M_K}\xrightarrow{}M_K$, i.e. architecture $f^{R_K}$ of resulting compressed model is determined at the first (and the only) compression step.


If we instead train the architecture $f^{R_K}$ from scratch, it is often impossible to achieve the quality comparable to the initial model $M$. If non-iterative approach is applied and we compress $M$ directly to the model from $\mathcal{M}_{R_K}$ (as it is done in  \cite{lebedev2014speeding, kim2015compression}), after fine-tunig we end up with a good baseline.
We show that our iterative approach (Algorithm~\ref{alg:compression}) beats the baseline in terms of compression ratio while preserving accuracy. (Section~\ref{sec:experiments}).

{\bf We can also describe our approach in a different way} taking into account that
each model from ${\mathcal{M}_{R_k}}$, ${k=1,\dots,K}$, can be mapped onto a model from $\mathcal{M}$ using the  operator $\mathcal{F}_{\mathrm{full}}$ (Figures~\ref{img:compress},\ref{img:compress_further}). Namely, for the initial architecture $f$, we iteratively reduce the set of valid parameters, $\Theta \supset \Theta^{R_1} \supset \dots \supset \Theta^{R_K}$,  and we search for the best parameters values  under the  imposed constraints.

Thus, every time we approximate weights (i.e. project weights to the smaller parameter subspace), we make a step away from the local minimum on the loss surface. Due to the continuity of the model, the more we reduce the weight rank, the bigger is the step and, hence, the more difficult it is to get back to a local optimum during the subsequent fine-tuning (because of the non-convexity of the optimization problem). 
In the iterative approach, in contrast to the non-iterative, the ranks decrease smoothly and gradually, and that allows to obtain a higher model compression rate with the same quality drop.

Moreover, our approach allows to perform compression by automatically searching for the best rank values $R_k, k=1,...,K$ (see Section~\ref{sec:rank_selection} for the details).

\section{Compression algorithm} \label{sec:compression_algo}

In this section, we formulate the optimization problems which need to be solved during one iteration of the compression algorithm, and we provide the detailed procedure for iterative neural network model compression. 
\subsection{One iteration of the algorithm}
{\bf To compress a layer} with a weight tensor $\theta$, for the first time, given rank $R$ we solve a problem of minimizing Frobenius norm, given rank~$R$:
\begin{equation} \label{eq:compress}
 \begin{gathered}
  \min_{\theta_{1}^{R},...,\theta_{N}^{R}}||\theta-  \hat{\theta}^R||,
  \quad \text{such that}  \\
  \mathcal{F}_{\mathrm{fact}}(\hat{\theta}^R) = \left(\theta_{1}^{R}, ..., \theta_{ N}^{R}\right),
 \end{gathered}
\end{equation}
where  $\theta_{1}^{R},\dots, \theta_{ N}^{R}$ denote components in a factorized form of the tensor and define weights of $N$ layers into which initial layer is decomposed during the ${\text{rank-}R}$ factorization
\footnote{Factors from tensor decomposition are equal to weights of decomposed layer up to the reshape operation.}.




{\bf To further compress an already decomposed layer}, we update its fine-tuned weights $\{\theta_{n}^{R}\}_{n = 1}^N$. Namely, given rank ${R' < R}$, we solve the following minimization problem:
\begin{equation} \label{eq:recompress}
\begin{gathered}
   \min_{\theta_{1}^{R'},...,\theta_{N}^{R'}}||\mathcal{F}_{\mathrm{full}}(\theta_{\mathrm{fact}}) -
   \hat{\theta}^{R'}||,
     \quad \text{such that}  \\
     \theta_{\mathrm{fact}} = (\theta_{1}^{R}, ..., \theta_{ N}^{R}),\\
     \mathcal{F}_{\mathrm{fact}}(\hat{\theta}^{R'}) = (\theta_{ 1}^{R'}, ..., \theta_{N}^{R'}),\\
\end{gathered}
\end{equation}
where $\theta_{1}^{R'},\dots, \theta_{ N}^{R'}$  denote updated weights (factor matrices) of the decomposed layer.

{\bf During the fine-tuning step} we minimize the loss function~$\ell$ given training data $\{(X_j, Y_j)\}_{j = 1}^J$, where $X_j$ is an input sample and $Y_j$ is a corresponding target value. Thus, we solve the following optimization problem:
\begin{equation}
     \mathcal{L}(\theta) \xrightarrow{} \min_{\theta \in \Theta^R_{\mathrm{fact}}},\quad \text{s.t.}\quad \mathcal{L}(\theta) = \sum_{j = 1}^J\ell\left(f^R(X_j, \theta), Y_j\right),
\end{equation}
where $f^R$ is our compressed architecture and $\Theta^R_{\mathrm{fact}}$ is the set of all possible model parameters.

\subsection{Iterative procedure}
Our proposed algorithm is an alternation of compression and fine-tuning steps with automatically selected ranks for the weights approximation (see Algorithm~\ref{alg:compression} for the details). 

At the compression step of each iteration for each of the selected layers, we solve the optimization problem \eqref{eq:compress} if a layer has not been compressed yet, and the optimization problem \eqref{eq:recompress} otherwise. The fine-tuning step is the same for all iterations.

\begin{algorithm}[H]
\caption{Iterative low-rank approximation algorithm for automated network compression}
\label{alg:compression}
\hspace*{\algorithmicindent} \textbf{Input:} Pre-trained original model, $M$ \\
\hspace*{\algorithmicindent} \textbf{Output:} Fine-tuned compressed model, $M^*$.
\begin{algorithmic}[1]
\State $M^* \gets M$
\While{desired compression rate is not attained or automatically selected ranks have not stabilized}
    \Statex
    \State $R \gets$ automatically selected ranks for low-rank tensor approximations of
        convolutional and fully-connected weight tensors.
    \Statex
    \State $\reallywidehat{M} \gets$ (further) compressed model obtained from $M$ by replacing layer weights with their rank-$R$ tensor approximations.
    \Statex
    \State $M^* \gets$ fine-tuned model $\reallywidehat{M}$.
\EndWhile
\end{algorithmic}
\end{algorithm}

\section{Layer compression in details} \label{sec:compress_one}

In this paper, we focus on Tucker and HOSVD (High Order Singular Value Decomposition) decompositions \cite{tucker1963implications}. 
Our framework can be used for other decompositions as well, and supplementary results using CP\footnote{CP is a canonical polyadic decomposition \cite{harshman1970foundations, hillar2013most}}-based  and SVD-based compressions are given in Appendix.

Since the definition of a tensor rank is not unified, we use a multilinear rank for Tucker (see definition below) 
and a CPD tensor rank for CP (Appendix). 



A \emph{Tucker decomposition} of an $N$-dimensional tensor  is a factorization into a small size \emph{core tensor} and $N$ \emph{factor matrices}. For a convolutional kernel $\theta \in \mathbb{R}^{d\times d\times C_{\mathrm{in}} \times C_{\mathrm{out}}}$,   with $C_{\mathrm{in}}$ input channels, $C_{\mathrm{out}}$ output channels, and a $d\times d$  spacial filter, it can be written as
\begin{equation} \label{eq:tucker_decomposition}
    \theta \approx \theta_C \times_{\mathrm{h}}  \theta_{\mathrm{h}}  \times_{\mathrm{w}}  \theta_{\mathrm{w}}  \times_{\mathrm{in}} \theta_{{\mathrm{in}}}  \times_{{\mathrm{out}}}  \theta_{{\mathrm{out}}},
\end{equation}
where $\theta_C$ is a $4$-dimensional core tensor, $\theta_{\mathrm{h}}, \theta_{\mathrm{w}}, \theta_{\mathrm{in}}, \theta_{\mathrm{out}}$ are matrices to be multiplied along each dimension of the core tensor. Symbols $\times_{\mathrm{h}}$, $\times_{\mathrm{w}}$ and $\times_{\mathrm{in}}, \times_{\mathrm{out}}$ denote multilinear products along spacial and channel dimensions respectively. 

If decomposition~\eqref{eq:tucker_decomposition} holds exatcly, the \textit{multilinear rank} of the tensor $\theta$ is defined as a tuple $(R_{\mathrm{h}}, R_{\mathrm{w}}, R_{\mathrm{in}}, R_{\mathrm{out}})$, where the {$n$-th} element, $n = 1 \dots 4$, is a rank of the \emph{dimension-$n$ unfolding} of the tensor\footnote{The dimension-$n$ unfolding of an $N$-dimensional tensor of size ${d_1\times...\times d_N}$ reorders the elements of the tensor into a matrix with $d_n$ rows and $d_1\dots d_{n-1}d_{n+1}\dots d_N$ columns}.

In convolutional kernels spacial dimensions usually are quite small. Therefore, similar to \cite{kim2015compression}, we factorize only two channel related dimensions, i.e. apply {\it Tucker-2 decomposition}, which is a specific form of the Tensor Train decomposition~\cite{oseledets2011tensor}:
\begin{equation} \label{eq:tucker2_decomposition}
    \theta \approx \theta_C \times_{\mathrm{in}} \theta_{{\mathrm{in}}} \times_{{\mathrm{out}}} \theta_{{\mathrm{out}}}.
\end{equation}
The corresponding multilinear rank equals $(d, d, R_{\mathrm{in}}, R_{\mathrm{out}})$, everywhere later we refer to it as $(R_{\mathrm{in}}, R_{\mathrm{out}})$.

\FloatBarrier
\subsection{First-time layer compression using Tucker-2 kernel approximation}

Let $\hat{\theta} \in \mathbb{R}^{d\times d\times C_{\mathrm{out}}\times C_{\mathrm{in}}}$ be a kernel approximation obtained via \text{Tucker-2} decomposition with rank $R = (R_{\mathrm{out}}, R_{\mathrm{in}})$. Since decomposition methods search directly for a factorized representation, we inroduce an \emph{operator $\mathcal{F}_{\mathrm{dec}}$}, which performs rank-$R$ approximation and factorization simultaneously, i.e.
\begin{equation}
    \mathcal{F}_{\mathrm{dec}}(\theta) = \mathcal{F}_{\mathrm{fact}}(\hat{\theta}) = (\theta_{\mathrm{C}}, \theta_{\mathrm{out}}, \theta_{\mathrm{in}}),
\end{equation}
where ${\theta_{\mathrm{out}}\in \mathbb{R}^{C_{\mathrm{out}}\times R_{\mathrm{out}}}}$,  ${\theta_{\mathrm{in}}\in \mathbb{R}^{C_{\mathrm{in}}\times R_{\mathrm{in}}}}$ are factor matrices, and $\theta_C \in\mathbb{R}^{d \times d\times R_{\mathrm{out}} \times R_{\mathrm{in}}}$ is a core tensor.

An output tensor $Y \in \mathbb{R}^{H' \times W' \times C_{\mathrm{out}}}$ given a layer input  $X \in \mathbb{R}^{ H \times W \times C_{\mathrm{in}}}$   can be calculated in a consecutive way as follows \cite{kim2015compression}
\begin{equation}
    Z_1 = \theta_{\mathrm{in}} * X, \quad Z_2 = \theta_{C} * Z_1, \quad Y = \theta_{\mathrm{out}} * Z_2,
\end{equation}
where operation $*$ denotes a convolution over all common dimensions.



\begin{figure}[h!]
\centering
\begin{tikzpicture}[scale=.35]
	\tikzcuboid{dimx=1, dimy=1, dimz=1, rotation=-15, angley=105, shifty=8, shiftx=30, scalex=2.4, scalez=2.4, scaley=0.7, 
	front/.style={fill=gray!40!white},
    top/.style={fill=gray!25!white},
    right/.style={fill=white!50!black}};
	\node at (.15, -1.9) {$C_{\mathrm{in}}$};
	\node at (3., -1.55) {$C_{\mathrm{out}}$};
	\node at (4.0, .3) {$d^2$};
	
	\node at (5., -.3) {$\simeq$};
	
	\tikzcuboid{dimx=1, dimy=1, dimz=1, shifty=-190, shiftx=38, scalex=1.0, scalez=2.0, scaley=1.0};
	\node at (-0.1, -8.3) {$d$};
	\node at (-0.8, -7.2) {$d$};
	\node at (.15, -5.4) {$C_{\mathrm{in}}$};
	\node at (3.7, -6.5) {$\small{@{C_{\mathrm{out}}}}$};
	
	\node at (5.5, -5.5) {\Large$\Rightarrow$};
	
	\node (A) at (1.3, -6.0) {};
	\node (B) at (1.3, -3.8) {};
	\node (C) at (1.3, -3.4) {};
	\node (D) at (1.3, -1.4) {};
	\draw[->, to path={-- (\tikztotarget)}, line width=1.3pt, dashed]
	(C) edge (D) ;
	\draw[-, to path={-- (\tikztotarget)}, line width=1.3pt, dashed] (A) edge (B);
	
	\node at (1.3, -3.8) {\small{Reshape}};
	
	\def\liney{-11.5}
	\node (X) at (-1.0, \liney) {X};
	\node (Y) at (4.2, \liney) {Y};
	\draw[->, to path={-- (\tikztotarget)}, line width=1.pt]
	(X) edge (Y) ;
	
	\draw[line width=.5pt] 
	    (7.1, -1.3) node{}
	-- (8.5, -1.3) node{}
	-- (8.5, .45) node{}
	-- (7.1, .45) node{}
	-- cycle;
	
	\tikzcuboid{dimx=1, dimy=1, dimz=1, rotation=-15, scalex=1.7, scalez=1.7, scaley=0.9, angley=105, shifty=-10, shiftx=357,  
	front/.style={fill=white},
    top/.style={fill=white},
    right/.style={fill=white}};
	
	\draw[line width=.5pt] 
	(18.0, -1.3) node{}
	-- (19.4, -1.3) node{}
	-- (19.4, .45) node{}
	-- (18.0, .45) node{}
	-- cycle;

	\tikzcuboid{dimx=1, dimy=1, dimz=1, rotation=-15, scalex=1, scaley=1.0, scalez=2, angley=105, shifty=-250, shiftx=232,};
	\tikzcuboid{dimx=1, dimy=1, dimz=1, rotation=-15, scalex=1, scaley=1.0, scalez=2, angley=105, shifty=-170, shiftx=375};
	\tikzcuboid{dimx=1, dimy=1, dimz=1, rotation=-15, scalex=1, scaley=1.0, scalez=2, angley=105, shifty=-250, shiftx=535};	
	
	\node (A) at (7.85, -1.2) {};
	\node (B) at (7.85, -8.2) {};
	\draw[->, to path={-- (\tikztotarget)}, line width=1.3pt, dashed]
	(A) edge (B);
	
	\node (C) at (18.7, -1.2) {};
	\node (D) at (18.7, -8.2) {};
	\draw[->, to path={-- (\tikztotarget)}, line width=1.3pt, dashed]
	(C) edge (D);

	\node (E) at (12.67, -1.2) {};
	\node (F) at (12.67, -3.0) {};
	\node (I) at (12.67, -3.2) {};
	\node (G) at (12.67, -5.4) {};
	\draw[-, to path={-- (\tikztotarget)}, line width=1.3pt, dashed] (E) edge (F);	
	\draw[->, to path={-- (\tikztotarget)}, line width=1.3pt, dashed] (I) edge (G);
	\node at (12.67, -3.1) {\small{Reshape}};
	
	\node (Xx) at (5.4, \liney) {X};
	\node (Z1) at (11.0, \liney) {$Z_1$};
	\node (Z2) at (16.0, \liney) {$Z_2$};
	\node (Yy) at (22.0, \liney) {Y};
	\draw[->, to path={-- (\tikztotarget)}, line width=1.pt]
	(Xx) edge (Z1) (Z1) edge (Z2) (Z2) edge (Yy) ;
	
	\node at (7.85, 1.05) {$C_{\mathrm{in}}$};
	\node at (6.3, -0.3) {$R_{\mathrm{in}}$};
	
	\node at (18.7, 1.05) {$R_{\mathrm{out}}$};
	\node at (20.6, -.3) {$C_{\mathrm{out}}$};
	
	\node at (9.8, -.3) {$\times_{\mathrm{in}}$};
	\node at (16.7, -.3) {$\times_{\mathrm{out}}$};
	
	\node at (11.5, 1.) {$R_{\mathrm{out}}$};
	\node at (13.4, 1.05) {$R_{\mathrm{in}}$};
	\node at (14.9, -.3) {$d^2$};
	
	\node at (6.8, -10.4) {$1$};
	\node at (6.1, -9.2) {$1$};
	\node at (6.8, -7.5) {$C_{\mathrm{in}}$};
	\node at (10.3, -8.6) {$\small{@{R_{\mathrm{in}}}}$};
	
	\node at (11.8, -7.6) {$d$};
	\node at (11.0, -6.4) {$d$};
	\node at (11.45, -4.7) {$R_{\mathrm{in}}$};
	\node at (15.6, -5.7) {$\small{@{R_{\mathrm{out}}}}$};
	
	\node at (17.4, -10.4) {$1$};
	\node at (16.7, -9.2) {$1$};
	\node at (17.1, -7.5) {$R_{\mathrm{out}}$};
	\node at (21.2, -8.6) {$\small{@{C_{\mathrm{out}}}}$};
\end{tikzpicture} 
\caption{Decomposition of a convolutional layer into 3 new ones using Tucker-2 kernel approximation. The top row shows an approximation of a 3D weight tensor with a low-rank tensor, which can be represented in Tucker-2 format ($\times_{\mathrm{in}}, \times_{\mathrm{out}}$ denote multilinear products along channel dimensions). The bottom row depicts how the initial layer is replaced with a sequence of 3 convolutional layers. Weights of new layers are the reshaped components of the factorized format for the low-rank tensor. The notation $@C_{\mathrm{out}}$ means that the 4D weight tensor has $C_{\mathrm{out}}$ output channels.} \label{fig:tucker2_compress}
\end{figure}
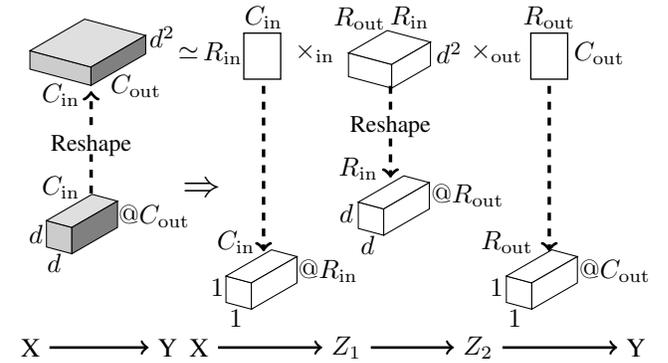

Therefore, the initial convolutional layer with kernel $\theta$ can be replaced by three convolutional layers (Figure~\ref{fig:tucker2_compress}). 
Indeed, we obtain $Z_1, Z_2, Y$ by sequentially propagating $X$ through the  layers with convolutions of spacial sizes $1\times 1$, $d\times d$ and $1\times 1$  respectively. 
Thus, for the decomposed layer we get $O(C_{\mathrm{in}}R_{\mathrm{in}} + d^2R_{\mathrm{out}}R_{\mathrm{in}} + C_{\mathrm{out}}R_{\mathrm{out}})$ layer parameters, and  propogation through this layer requires $O(HWC_{\mathrm{in}}R_{\mathrm{in}} + H'W'(d^2R_{\mathrm{out}}R_{\mathrm{in}} + C_{\mathrm{out}}R_{\mathrm{out}}))$  operations.
\FloatBarrier
\noindent



\subsection{Further compression of a Tucker-2 decomposed layer}
To perform further compression we need to update  weights $\theta_{\mathrm{fact}} = (\theta_{\mathrm{C}}, \theta_{\mathrm{out}}, \theta_{\mathrm{in}})$ of the decomposed layer, i.e. to find $\hat{\theta}'_{\mathrm{fact}} = (\theta'_{\mathrm{C}}, \theta'_{\mathrm{out}}, \theta'_{\mathrm{in}})$ such that $\mathcal{F}_{\mathrm{full}}(\hat{\theta}'_{\mathrm{fact}})$ has multilinear rank  $R' = (R'_{\mathrm{out}}, R'_{\mathrm{in}})$, $R' \le R$ (elementwise comparison).



The  naive way to do that is to obtain a new core and factor matrices by approximating  tensor $\mathcal{F}_{\mathrm{full}}(\theta_{\mathrm{fact}})$ with Tucker-2 decomposition (the path along dashed arrows in Figure~\ref{img:compress_further}). 

\begin{figure}[h!]
    \centering
    \begin{tikzpicture}[scale=.35]
        \draw[line width=.5pt] 
    	(-0.5, 0) node{}
    	-- (1.5, 0.0) node{}
    	-- (1.5, 2.5) node{}
    	-- (-0.5, 2.5) node{}
    	-- cycle;
    	\node at (-1.4, 1.25) {$C_{\mathrm{in}}$};
    	\node at (0.5, 3.2) {$R_{\mathrm{in}}$};	
    	\node at (0.5, 1.25) {$\theta_{\mathrm{in}}$};	
    	
    	\node at (2.5, 1.25) {$\times_{\mathrm{in}}$};	
    	
    	\tikzcuboid{dimx=1, dimy=1, dimz=1, rotation=-15, scalex=1.9, scalez=1.9, scaley=1.0, angley=105, shiftx=150, shifty=35};	
    	\node at (5.4, 1.6) {$\theta_{\mathrm{C}}$};
    	\node at (4.1, 2.5) {$R_{\mathrm{in}}$};
    	\node at (6.7, 2.5) {$R_{\mathrm{out}}$};	
    	
    	\node at (8.3, 1.25) {$\times_{\mathrm{out}}$};	
    	
    	\draw[line width=.5pt] 
    	(9.5, 0) node{}
    	-- (11.5, 0.0) node{}
    	-- (11.5, 2.5) node{}
    	-- (9.5, 2.5) node{}
    	-- cycle;
    	\node at (10.5, 1.25) {$\theta_{\mathrm{out}}$};		
    	\node at (12.8, 1.25) {$C_{\mathrm{out}}$};
    	\node at (10.5, 3.2) {$R_{\mathrm{out}}$};	
    	
    	\node at (16.5, 2.5) {\small\text{weights update}};
    	\node at (16.5, 1.25) {\Large$\simeq$};	
    	
    	\node at (4.3, -3.75) {\Large$\simeq$};	
    	\draw[line width=.5pt] 
    	(6.8, -5.0) node{}
    	-- (8.4, -5.0) node{}
    	-- (8.4, -2.5) node{}
    	-- (6.8, -2.5) node{}
    	-- cycle;
    	\node at (6.0, -3.75) {$C_{\mathrm{in}}$};
    	\node at (7.6, -1.8) {$R'_{\mathrm{in}}$};	
    	\node at (7.6, -3.75) {$\theta'_{\mathrm{in}}$};	
    	
    	\node at (9.4, -3.75) {$\times_{\mathrm{in}}$};

    	\tikzcuboid{dimx=1, dimy=1, dimz=1, rotation=-15,  scalex=2.0, scalez=1.8, scaley=1.1, angley=105, shiftx=340, shifty=-103};	
    	\node at (12.1, -3.2) {$\theta^*_{\mathrm{C}}$};
    	\node at (11.3, -5.5) {$R'_{\mathrm{in}}$};	
    	\node at (13.8, -5.2) {$R'_{\mathrm{out}}$};

        \node at (15.1, -3.75) {$\times_{\mathrm{out}}$};	
        	
        \draw[line width=.5pt] 
        (16.3, -5.0) node{}
        -- (18.2, -5.0) node{}
        -- (18.2, -2.5) node{}
        -- (16.3, -2.5) node{}
        -- cycle;
    \node at (19.5, -3.75) {$C_{\mathrm{out}}$};
    \node at (17.3, -1.8) {$R'_{\mathrm{out}}$};
    \node at (17.3, -3.75) {$\theta'_{\mathrm{out}}$};

    \draw [decorate,decoration={brace, amplitude=10pt}, xshift=-0.5cm, yshift=0pt]
    (3.6, 2.6) -- (7.95, 2.6) node [black,midway,xshift=-0.6cm]{};

    \tikzcuboid{dimx=1, dimy=1, dimz=1, rotation=-15, scalex=2.0, scalez=1.9, scaley=0.5, angley=105, shiftx=140, shifty=220};	
	\node at (5.0, 7.5) {$\theta^*_{\mathrm{C}}$};
    \node at (4.2, 5.8) {$R'_{\mathrm{in}}$};
    \node at (6.9, 6.1) {$R'_{\mathrm{out}}$};	
    	
    \node at (2.3, 7.5) {$\times_{\mathrm{in}}$};	
    \node at (8.1, 7.5) {$\times_{\mathrm{out}}$};		
    	
    \draw[line width=.5pt] 
    (-1.0, 6.35) node{}
    -- (1.0, 6.35) node{}
    -- (1.0, 8.6) node{}
    -- (-1.0, 8.6) node{}
    -- cycle;
    	
    \node at (0.0, 7.475) {$\theta^*_{\mathrm{in}}$};
    \node at (-2.0, 7.475) {$R_{\mathrm{in}}$};
    \node at (0.0, 9.3) {$R'_{\mathrm{in}}$};	
    	
    \draw[line width=.5pt] 
    (9.5, 6.35) node{}
    -- (11.5, 6.35) node{}
    -- (11.5, 8.6) node{}
    -- (9.5, 8.6) node{}
    -- cycle;
    	
    \node (C) at (5.25, 3.1) {};
    \node (D) at (5.25, 6.7) {};
    \draw[->, to path={-- (\tikztotarget)}, line width=1.pt, dashed]
    	(C) edge (D);
    	
    \node at (10.5, 7.475) {$\theta^*_{\mathrm{out}}$};
    \node at (12.7, 7.475) {$R_{\mathrm{out}}$};
    \node at (10.5, 9.3) {$R'_{\mathrm{out}}$};	
    
    \end{tikzpicture}
    \caption{A low-rank approximation of a 3D tensor represented in Tucker-2 format.} \label{fig:tucker2_compress_further}
\end{figure}
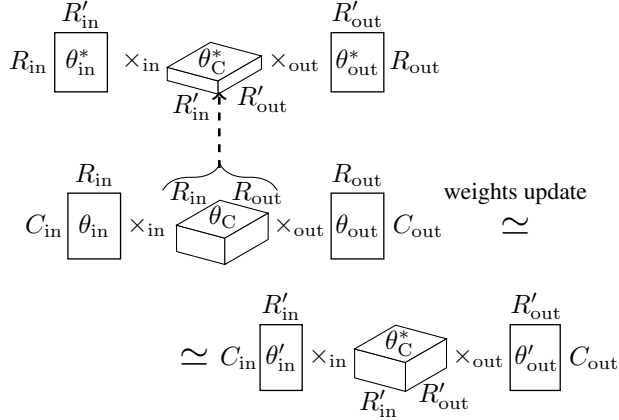

We propose to use a more efficient update based on the properties of Tucker decomposition (Figure~\ref{fig:tucker2_compress_further}). Namely, we  approximate the core $\theta_{\mathrm{C}}$ using Tucker-2 and then update the weights in the following way
\begin{equation}
\begin{gathered}
 \mathcal{F}_{\mathrm{dec}}(\theta_C) = (\theta^*_{\mathrm{C}}, \theta^*_{\mathrm{out}}, \theta^*_{\mathrm{in}}),\\
  \theta'_{\mathrm{C}} =  \theta^*_{\mathrm{C}}, \quad
     \theta'_{\mathrm{in}} =  \theta_{\mathrm{in}}  \theta^*_{\mathrm{in}}, \quad\theta'_{\mathrm{out}} = \theta_{\mathrm{out}}  \theta^*_{\mathrm{out}}. 
\end{gathered}
\end{equation}

\FloatBarrier
\noindent

\section{Automatic rank selection} \label{sec:rank_selection}
In Section~\ref{sec:problem_statement} we have shown that iterative architecture compression and model fine-tuning \eqref{eq:model_sequence} is equivalent to iterative  parameter set reduction ($\Theta \supset \Theta^{R_1} \supset\Theta^{R_2}\supset \dots$) for the initial architecture $f$ and fine-tuning with parameter constraints.

To automatically choose gradually decreasing ranks ${R_1 > R_2 > \dots}$ for low-rank  approximation of $\theta$,  we experiment with two different scenarios: Bayesian approach and constant compression rate. Bayesian approach based on rank estimation in order to remove redundancy from the weight tensor. Constant compression rate approach based on rank calculation in order to obtain the desired parameter reduction rate after several iterations. 




\subsection{Bayesian approach}
For convenience, we introduce two notations: an \textit{extreme rank} and a \textit{weakened rank}. The {extreme rank} is the value at which almost all redundancy is eliminated from the tensor after decomposition. The {weakened rank} is the value at which a certain amount of redundancy is preserved in the tensor after decomposition.

In Bayesian approach, firstly, extreme rank $R_{e}$ is found via \textit{GAS of EVBMF} (Global Analytic Solution of Empirical Variational Bayesian Matrix Factorization \cite{nakajima2012perfect}), and secondly, a rank weakening is performed. 

The GAS of EVBMF can automatically find matrix rank by performing Bayesian inference, however, it
provides a suboptimal solution.
Unlike authors of \cite{kim2015compression}, we use   GAS of EVBMF not to set a rank for a  weight approximation (i.e. ${R = R_{\mathrm{evbmf}}}$), but only to determine extreme rank (i.e. ${R_{\mathrm{extr}} = R_{\mathrm{evbmf}}}$)\footnote{In \cite{kim2015compression} instead of  EVBMF  they use non-empirical version of VBMF, which requires manual setting of additional global parameter.}.


To determine the extreme rank for Tucker-2 approximation using the GAS of EVBMF, we apply it to the unfoldings of the weight tensor associated with channel dimensions \cite{kim2015compression}. That is, at the $(k+1)$-th iteration   we apply it to matrices  of sizes ${R^{k}_{\mathrm{in}} \times d^2R^{k}_{\mathrm{out}}}$ and $ {R^{k}_{\mathrm{out}} \times d^2R^{k}_{\mathrm{in}}}$.

The weakened rank $R_{\mathrm{weak}}$ depends linearly on the extreme rank and serves to preserve more redundancy in the decomposed tensor. Setting $R = R_{\mathrm{weak}}$ facilitates fine-tuning and yields a compression step with better accuracy.

The weakened rank is defined as follows: 
\begin{equation}
    R_{\mathrm{weak}} = R_{\mathrm{init}} -  w (R_{\mathrm{init}} - R_{\mathrm{extr}}),
\end{equation}
where $w$, is a hyperparameter called \emph{weakening factor}, ${0 < w < 1}$, and $R_{\mathrm{init}}$ stands for initial rank. This results in $R_{\mathrm{extr}} \le R_{\mathrm{weak}} \le R_{\mathrm{init}}$.
Our experiments show that the optimal value for $w$ is in the range: $0.5 \leq w\leq 0.9$. If the initial rank is less than $21$, our algorithm considers such kernels as small enough and does not compress them.

\subsection{Constant compression rate}
Ranks for tensor approximations can be chosen based on parameter reduction rate that we want to achieve at each compression step. By choosing rank in such a way, we can control the speed-up of each convolutional layer. 

Suppose we want to reduce the number of kernel  parameters in initial convolutional layer, $d^2C_{\mathrm{in}}C_{\mathrm{out}}$ times $\alpha$. In Tucker-2 case, having $R_{\mathrm{in}}C_{\mathrm{in}} + R_{\mathrm{out}}R_{\mathrm{in}}d^2 + R_{\mathrm{out}}C_{\mathrm{out}}$ parameters in the decomposed layer, and assuming the multilinear rank has the form $(\beta R, R)$,\ $\beta > 0$, we can derive
\begin{equation} \label{eq:rank_tucker}
\begin{gathered}
 R \le \frac{-\frac{C_{\mathrm{in}}+\beta C_{\mathrm{out}}}{\beta d^2} + \sqrt{\frac{(C_{\mathrm{in}}+\beta C_{\mathrm{out}})^2}{\beta^2d^4} + \frac{4C_{\mathrm{in}}C_{\mathrm{out}}}{\beta \alpha}}}{2}.
\end{gathered}
\end{equation}



Therefore, to achieve times $\alpha$ parameters reduction using Tucker-2 tensor approximation, we choose ranks according to the inequality \eqref{eq:rank_tucker}.

\section{Experiments} \label{sec:experiments}
In this paper, we focus on compressing object detection neural network models. 
We apply our {\bf Multi-Stage COmpression  (MUSCO)} approach on  Faster R-CNN~\cite{fpn}  and Faster R-CNN \cite{ren2015faster} with ResNet-50 and VGG-16 backbones.  To show the applicability of our method to the variety of tasks, we apply it to compress several classifications and other object detection benchmarks.   We have made the code publicly available\footnote{https://github.com/juliagusak/musco}.

In this section, we report results on  model compression obtained by MUSCO based on Tucker-2 decomposition with an automatic rank selection. To one iteration of MUSCO we further refer as {\emph{Tucker2-iter}}. 


\subsection{Compression of Faster R-CNN}
This section demonstrates the effectiveness of MUSCO algorithm in compressing Faster R-CNN with VGG-16 backbone\footnote{https://github.com/chenyuntc/simple-faster-rcnn-pytorch}, Faster R-CNN C4 (used for PASCAL VOC dataset) and Faster R-CNN FPN (used for COCO dataset) with ResNet-50 backbone\footnote{https://github.com/facebookresearch/maskrcnn-benchmark}. In our experiments, we focus on backbone compression since tensor methods can reduce the parameter redundancy, which usually occurs in convolutional layers. The quality of object detection tasks is evaluated using mAP (mean Average Precision) metrics.

\subsubsection{Faster R-CNN with VGG-16 backbone}

Faster R-CNN with VGG-16 backbone has been fine-tuned and evaluated on  {Pascal VOC 2007~\cite{voc}} train and test datasets respectively.  

At the bottom part of Table~\ref{table:vggvoc} we provide compressed models obtained via several compression iterations along with models compressed at one step. Model parameter {\bf nx} means that we select ranks for Tucker-2 decomposition based on constant compression rate strategy.  For example,  {MUSCO(nx, 3.16, 2)} is a compressed model obtained after two compression steps using $3.16\times$ parameter reduction at each step. 

You can see from the table that the iterative approach allows obtaining higher FLOPs reduction than non-iterative one at a similar mAP level. For example, in this sense {MUSCO(nx, 1.77, 2)} and {MUSCO(nx, 2, 2)} outperform {Tucker2-iter (nx, 3.16)}, {MUSCO(nx, 3.16, 2)} and {MUSCO(nx, 1.77, 4)} is better than {Tucker2 (nx, 10)}.

Comparing to the state-of-the-art methods from Table~\ref{table:vggvoc},  the MUSCO approach outperforms all models except the one from ~\cite{He_2018_ECCV}. However, our model MUSCO(nx, 3.16, 2) gives $10.49\times$ FLOPs reduction comparing to $4\times$ from~\cite{He_2018_ECCV} and outperforms the latter in terms of absolute mAP value.
The FLOPS are computed in the same way as in \cite{He_2017_ICCV}.

\begin{table}[!h]
\begin{center}
\begin{tabular}{lcc}
\hline
Model & FLOPs & mAP\\
\hline
\multicolumn{3}{c}{FASTER R-CNN (VGG-16) @ VOC2007} \\
\hline
{\cite{He_2017_ICCV} baseline} & $1\times$ & 68.7 \\ \hline
\textit{Channel Prunning} \cite{He_2017_ICCV} &  $~4\times$  & 66.9(-1.8) \\ 
\textit{Accelerating VD} \cite{zhang2016accelerating} & $~4\times$  & 67.8(-0.9) \\ 
\textit{AutoML Compression} \cite{He_2018_ECCV} & $4\times$  &  68.8(+0.1) \\ \hline
{Used baseline} & $1.0\times$ & {71.1} \\ \hline
Tucker2-iter (nx, 3.16) &  $3.16\times$  & 70.7(-0.4) \\ 
\textbf{MUSCO(nx, 1.77, 2)} &  $\mathbf{3.72\times}$  & \textbf{70.4(-0.7)}\\
\textbf{MUSCO(nx, 2, 2)} &  $\mathbf{4.69\times}$  & \textbf{70.1(-1.0) }\\ 
Tucker2-iter (nx, 10) &  $9.67\times$  & 68.6(-2.5) \\ 
\textbf{MUSCO(nx, 3.16, 2)} &  $\mathbf{10.49\times}$  & \textbf{69.2(-1.9)} \\ 
\textbf{MUSCO(nx, 1.77, 4)} &  $\mathbf{13.95\times}$  & \textbf{68.3(-2.8)} \\ 
\hline
\end{tabular}
\end{center}
\caption{Comparison of Faster R-CNN (with VGG-16 backbone) compressed models on VOC2007 evaluation dataset. {MUSCO(nx, 3.16, 2)} is a compressed model obtained after two compression steps using $3.16\times$ parameter reduction  at each step
}
\label{table:vggvoc}
\end{table}








\subsubsection{Faster R-CNN  with ResNet-50 backbone}
\label{sec:coco}
To the best of our knowledge, there are currently no sufficiently good compression methods for ResNet50-based Faster R-CNN and lack of methods, evaluated on COCO~\cite{coco} dataset, while both aspects are currently standard for the object detection research area. 

In Tables ~\ref{table:resnetvoc} and \ref{table:coco} we provide compression results obtained by MUSCO  based on Tucker-2 decomposition for Faster R-CNN C4 and Faster R-CNN with ResNet-50 backbone.

Faster R-CNN has been fine-tuned and evaluated on {COCO-2014} dataset. Faster R-CNN with ResNet-50 backbone has been fine-tuned on Pascal VOC 2007+2012 train datasets and evaluated on Pascal VOC 2007 test dataset. 

On Pascal VOC2007 our MUSCO models achive $1.39\times$ and $1.57\times$ FLOPs reduction with mAP increase  by 2.0 and 0.4, respectively (Table~\ref{table:resnetvoc}).

\begin{table}[!h]
\begin{center}
\begin{tabular}{lcc}
\hline
Model &  FLOPs & mAP\\\hline
\multicolumn{3}{c}{FASTER R-CNN C4 (RESNET-50) @ VOC2007} \\\hline
{Used baseline} & $1.0\times$ & {75.0} \\ \hline
Tucker2-iter (nx, 1.4)&  $1.17\times$  & 76.8(+1.8) \\ 
\textbf{MUSCO(nx, 1.4, 2)}&  $\mathbf{1.39\times}$  & \textbf{77.0(+2.0)} \\ \hline
\textbf{MUSCO(nx, 1.4, 3)}&  $\mathbf{1.57\times}$  & \textbf{75.4(+0.4)} \\ 
Tucker2-iter (nx, 3.16)\hspace{-10pt}&  $1.49\times$  & 75.0(+0.0) \\ 
\hline
\end{tabular}
\end{center}
\caption{Comparison of Faster R-CNN (with ResNet-50 backbone)  compressed models on Pascal VOC2007 evaluation dataset.
}
\label{table:resnetvoc}
\end{table}

On COCO2014 dataset our different MUSCO models outperform one-step compression (Table \ref{table:coco}).  

\begin{table}[!h]
\begin{tabular*}{0.5\textwidth}{lccc}
\hline
Model & FLOPs & mAP & mAP.50\\
 \hline
 \multicolumn{4}{c}{FASTER R-CNN FPN (RESNET-50) @ COCO2014} \\
\hline
Original & $1.0\times$ \hspace{-13pt} & 37.7 & 59.1 \\ \hline
Tucker2-iter(vbmf, 0.7)\hspace{-10pt} & $\mathbf{1.2\times}$ \hspace{-13pt} & \textbf{36.3(-1.4)} & \textbf{57.3(-1.8) }\\
\textbf{MUSCO(vbmf, 0.7, 2)}\hspace{-10pt} & $\mathbf{1.7\times}$\hspace{-13pt}  & \textbf{36.2(-1.5)} & \textbf{57.1(-2.0)} \\
\textbf{MUSCO(nx, 3, 4)} & $\mathbf{1.8\times}$\hspace{-13pt}  & \textbf{35.4(-2.3)} & \textbf{56.2(-2.9)} \\
Tucker2-iter(vbmf, 0.9)\hspace{-10pt} & $2.0\times$\hspace{-13pt}  & 33.8(-3.9) & 54.0(-5.1) \\
\hline 
\end{tabular*}
~
\caption{Comparisom of Faster R-CNN (with ResNet-50 backbone) compressed models on COCO2014 dataset. {MUSCO (vbmf, 0.7, 2)} corresponds to the two-iteration compression with automatically selected ranks using GAS of EVBMF and rank weakening with weakeinig factor equals 0.7. 
}
\label{table:coco}
\end{table}

\subsection{Compression of other models}

To demonstrate the difference between our approach and non-repetitive one, we performed compression using the GAS of EVBMF as the only rank selector. Results are shown in Table~\ref{table1}.

\begin{table}[!h]
\begin{center}
\begin{tabular}{lcc}
\hline
Model & MUSCO & Tucker2-iter \\ 
\hline
AlexNet & -0.81 & -4.2 \\
VGG-16 & -0.15 & -2.8 \\
YOLOv2 & -0.19 & -3.1 \\
Tiny YOLOv2 & -0.10 & -2.7 \\
\hline
\end{tabular}
\end{center}
~
\caption{Quality drop after iterative compression and one-time compression. For AlexNet and VGG-16 metric is $\Delta$ Top-5 accuracy, for YOLO - {$\Delta$ mAP}}
\label{table1}
\end{table}

Comparing the results of compression ratio and speed up in two tables, we can say that iterative approach showed itself better for almost all networks. We can also see that the highest speedup is achieved on CPUs as follows from Table~\ref{table2}. Moreover, we can see that speed up for different processor series is almost the same.
To achieve these results, fine-tuning has been performed as long as error rate decreased for both iterative approach and one-time compression approach. For AlexNet and VGG we measured Top-5 accuracy and for YOLOv2 and Tiny YOLO we measured mAP.




\begin{table}[!h]
\begin{center}
\begin{tabular}{lcccc}
\hline
Model & Size & CPU1 & CPU2 & GPU \\
\hline
AlexNet & $4.90\times$ & $4.73\times$ & $4.55\times$  & $2.11\times$ \\
VGG-16 & $1.51\times$ & $3.11\times$ & $3.23\times$ & $2.41\times$ \\
YOLOv2 & $2.13\times$ & $2.07\times$ & $2.16\times$ & $1.62\times$ \\
Tiny YOLOv2 & $2.30\times$ & $2.35\times$ & $2.28\times$ & $1.71\times$ \\
\hline
\end{tabular}
\end{center}
~
\caption{Results of iterative low-rank approximation for AlexNet, VGG-16, YOLOv2, and Tiny YOLO. These tests were performed on CPUs of two different series and on GPU: Intel Core i5-7600K, Intel Core i7-7700K and NVIDIA GeForce GTX 1080 Ti respectively.}
\label{table2}
\end{table}

\section{Related work}
\label{sec:related_work}
There are several works devoted to the deep convolutional neural networks compression. Authors of \cite{han2015deep} proposed a pipeline that consists of three different methods: pruning, trained quantization, and Huffman coding. They demonstrated the possibility of significantly reducing storage requirements by combining different techniques. Our method differs since we focus not only on compression ratio but also on speedup and seamless integration into any framework

Various methods based on quantization were been proposed in \cite{lin2016fixed, miyashita2016convolutional, park2017weighted}. The main goal of quantization is to reduce the number of bits required for weight storage. Our approach differs from that one because we compress networks by decomposing tensors and reducing ranks. It is worth mentioning that quantization can be used after applying our method and serve as an additional method of compression. But quantization may require altering a framework and a significant speedup can be achieved only taking into account the peculiarities of the hardware.

There are several experiments on training low precision networks \cite{baldassi2015subdominant, rastegari2016xnor}. Their methods allow to use only 2 bits for weight storage but accuracy is much lower than in full precision networks, and it is not a compression algorithm because such networks have to be trained from scratch.

Several approaches based on different algorithms of low-rank approximation were proposed in \cite{lebedev2014speeding, denton2014exploiting}. The authors of \cite{denton2014exploiting} have demonstrated the successful application of singular value decomposition (SVD) to fully connected layers. Further, the authors of \cite{lebedev2014speeding} found a way to decompose 4-dimensional convolutional kernel tensor by applying canonical polyadic (CP) decomposition. But these approaches are able to compress only one or a couple of layers. Moreover, for each layer the rank is unique and the process of rank selection has to be performed manually every time.

Another way to compress a whole network was introduced in \cite{kim2015compression}. The approach used in their work is automated. Authors combined two different decompositions to be able to compress both fully connected and convolutional layers. To compress fully connected layers they adopted the approach used in \cite{denton2014exploiting} and applied SVD. For convolutional layers, the authors applied a Tucker decomposition \cite{tucker1966some}. Unlike \cite{lebedev2014speeding, denton2014exploiting}, the authors have found a way to automatically select ranks without any manual search. Ranks are determined by a global analytic solution of variational Bayesian matrix factorization (VBMF) \cite{nakajima2012perfect}. We found that the global analytic VBMF provides ranks for which it is difficult to restore the initial accuracy by fine-tuning for deep networks. In our algorithm, we  use the global analytic EVBMF but to select the extremal ranks which will be weakened afterward.

CP decomposition which was used by \cite{lebedev2014speeding} is a special case of Tucker decomposition, where the core tensor is constrained to be superdiagonal. In our approach, we use Tucker-2 decomposition. To compress fully connected layers we adopt SVD as it was proposed in \cite{denton2014exploiting}.

Our approach is different from these methods because all of them apply decomposition algorithms only one time per layer, and ranks provided by the global analytic VBMF can be considered as upper bounds for compression. Our algorithm is iterative, and decomposition algorithms can be applied multiple times for the same layer. Moreover, we can achieve a higher compression rate because we do not have such boundary.





\section{Conclusion}
In this paper, we addressed the problem of compression of deep convolutional neural networks. We proposed a multi stage compression algorithm  MUSCO for neural network compression, which performs gradual redundancy reduction. Our method consists of two repetitive steps: compression and fine-tuning. Compression step  includes automatic rank  selection and low-rank tensor factorization according to the selected rank. We evaluated our approach on the following deep networks used for object detection: Faster R-CNN with VGG-16 and ResNet-50 backbones, YOLOv2, Tiny YOLO and classification: AlexNet, VGG-16. Experimental results show that our iterative approach outperforms non-repetitive ones in the compression ratio providing less accuracy drop.

Our method is designed to compress any neural network architecture with convolutional and fully connected layers using Tucker-2, CP or SVD decomposition with two different strategies of automatic rank selection.  As future work we plan to increase the variety of matrix/tensor decompositions used by MUSCO approach at the compression step. Also we will investigate the effect of combining 
our approach with hardware-dependent approaches such as quantization and channel pruning approaches.

{\small
\bibliographystyle{ieee}
\bibliography{musco.bib}
}
\newpage
\section{Appendix}
\subsection{Compression using CP decomposition}
CP decomposition (CPD) is a special case of Tucker decomposition when the cube core tensor has nonzero elements only on the main diagonal. Thus, for an ${N\text{-dimensional}}$ tensor CPD (we call it {\it CPD-$N$} hereinafter) is defined by $N$ factor matrices.

Unlike  \cite{lebedev2014speeding}, instead of applying CPD-4 to a {4-dimensional} kernel tensor $\theta$, we propose to apply CPD-3 to a reshaped tensor  of size $d^2\times C_{\mathrm{out}}\times C_{\mathrm{in}}$. The advantage of CPD-3 over CPD-4 is that it allows catching dependencies within spatial dimensions. Moreover, it yields faster convergence during approximation because of a decreased number of factors.
\subsubsection{First time layer compression using CP-format kernel approximation}

Using  CPD-3 decomposition with CP-rank $R$ (here \textit{CP-rank} is defined as the minimum number of rank-one tensors required to yield its exact CP decomposition), a factorized form of the target kernel approximation $\hat{\theta} \in \mathbb{R}^{ d^2 \times C_{\mathrm{out}}\times C_{\mathrm{in}} }$ is equal to 
\begin{equation}
     \mathcal{F}_{\mathrm{fact}}(\hat{\theta}) = (\theta_{\mathrm{d^2}}, \theta_{\mathrm{out}}, \theta_{\mathrm{in}}),      
\end{equation}
where  $\theta_{d^2}\in \mathbb{R}^{d^2\times R}$, $\theta_{\mathrm{out}}\in \mathbb{R}^{C_{\mathrm{out}}\times R}$,  $\theta_{\mathrm{in}}\in \mathbb{R}^{C_{\mathrm{in}}\times R}$ are factor matrices (components).


Similar to the Tucker-2 case, we replace one convolutional layer with a sequence of three layers when we apply CPD-3 to approximate a kernel tensor. The only difference is that the second convolutional layer is determined by grouped convolutions (with $R$ groups), not by a standard one.
Therefore, if $X \in \mathbb{R}^{ H \times W \times C_{\mathrm{in}}}$ and $Y \in \mathbb{R}^{H' \times W' \times C_{\mathrm{out}}}$ are layer input and output respectively, we have $O(R(C_{\mathrm{in}} + d^2 + C_{\mathrm{out}}))$  layer parameters, and the computational cost equals $O(R(HWC_{\mathrm{in}} + H'W'd^2R+H'W'C_{\mathrm{out}}))$.



\begin{figure}[h!]
    \centering
    \begin{tikzpicture}[scale=.35]
        \draw[line width=.5pt] 
        (-0.5, 0) node{}
        -- (1.5, 0.0) node{}
        -- (1.5, 2.5) node{}
        -- (-0.5, 2.5) node{}
        -- cycle;
        \node at (-1.4, 1.25) {$d^2$};
        \node at (0.5, 3.2) {$R$};    
        
        \node at (2.5, 1.25) {$\circ$};    
        
        \draw[line width=.5pt] 
        (3.5, 0) node{}
        -- (5.5, 0.0) node{}
        -- (5.5, 2.5) node{}
        -- (3.5, 2.5) node{}
        -- cycle;
        
        \node at (4.5, 3.2) {$R$};
        \node at (6.8, 1.25) {$C_{\mathrm{out}}$};
        
        \node at (8.3, 1.25) {$\circ$};    
        
        \draw[line width=.5pt] 
        (9.3, 0) node{}
        -- (11.3, 0.0) node{}
        -- (11.3, 2.5) node{}
        -- (9.3, 2.5) node{}
        -- cycle;
        \node at (12.4, 1.25) {$C_{\mathrm{in}}$};
        \node at (10.3, 3.2) {$R$};    
        
        \node at (16.5, 2.5) {\small\text{weights update}};
        \node at (16.5, 1.25) {\Large$\simeq$};    
        
        \node at (4.3, -3.75) {\Large$\simeq$};    
        \draw[line width=.5pt] 
        (6.8, -5.0) node{}
        -- (8.8, -5.0) node{}
        -- (8.8, -2.5) node{}
        -- (6.8, -2.5) node{}
        -- cycle;
        \node at (5.9, -3.75) {$d^2$};
        \node at (7.8, -1.8) {$R'$};    
        
        \node at (9.7, -3.75) {$\circ$};    
        
        \draw[line width=.5pt] 
        (10.6, -5.0) node{}
        -- (12.6, -5.0) node{}
        -- (12.6, -2.5) node{}
        -- (10.6, -2.5) node{}
        -- cycle;
        \node at (13.8, -3.75) {$C_{\mathrm{out}}$};
        \node at (11.5, -1.8) {$R'$};    
        
        \node at (14.9, -3.75) {$\circ$};    
            
        \draw[line width=.5pt] 
        (15.8, -5.0) node{}
        -- (17.8, -5.0) node{}
        -- (17.8, -2.5) node{}
        -- (15.8, -2.5) node{}
        -- cycle;
    \node at (19.0, -3.75) {$C_{\mathrm{in}}$};
    \node at (16.8, -1.8) {$R'$};
        
    \end{tikzpicture}
    \caption{A low-rank approximation of a 3D tensor represented in CPD-3-format. $\circ$ is the outer product operation.} 
\end{figure}

\subsubsection{Further compression of CP - decomposed layer}

For further compression we update  weights $\theta_{\mathrm{fact}} = (\theta_{\mathrm{d^2}}, \theta_{\mathrm{out}}, \theta_{\mathrm{in}})$ of the decomposed layer, i.e.  find $\hat{\theta}'_{\mathrm{fact}} = (\theta'_{\mathrm{d^2}}, \theta'_{\mathrm{out}}, \theta'_{\mathrm{in}})$, such that $\mathcal{F}_{\mathrm{full}}(\hat{\theta}'_{\mathrm{fact}})$ has CP-rank  $R' \le R$.

A naive way to update weights of a decomposed layer is to construct a full tensor from fine-tuned weights, approximate it with CPD-3 with reduced CP-rank and update the weights using new factors. However, that might not guarantee an appropriate approximation error. 

To improve the convergence, we approximate current fine-tuned weights with  CP factors of lower rank directly, using an ALS-type algorithm.


\subsubsection{Rank selection}
Suppose we want to reduce $\alpha$ times the number of kernel  parameters in initial convolutional layer, $d^2C_{\mathrm{in}}C_{\mathrm{out}}$.  Having $R(C_{\mathrm{in}}+d^2+C_{\mathrm{out}})$ kernel parameters in a {CPD-3} decomposed layer, the estimated CP-rank $R$ should satisfy the following inequality
\begin{equation} \label{eq:rank_cp}
\begin{gathered}
 R \le \frac{d^2C_{\mathrm{in}}C_{\mathrm{out}}}{\alpha (C_{\mathrm{in}}+d^2+C_{\mathrm{out}})}.
 \end{gathered}
\end{equation}
Therefore, to achive times $\alpha$ parameters reduction using CPD-3 tensor approximation, we choose ranks according to the inequality \eqref{eq:rank_cp}. For CP case, please,  see Appendix.

\subsubsection{Experiments}
We tested CPD-3 based compression on Faster R-CNN model with ResNet-50 backbone. Compressing a part of layers at each step, we  achive the following results 
\begin{table}[!htb].
\begin{tabular*}{0.49\textwidth}{lcc}
\hline
Model &  FLOPs & mAP\\\hline
\multicolumn{3}{c}{FASTER R-CNN C4 (RESNET-50) @ VOC2007} \\\hline
{Used baseline} & $1.0\times$ & {75.0} \\ \hline
\textbf{MUSCO} $\left(\text{nx}, 5, 3 \times 1/3\right)$ &  $1.63\times$  & 74.79 (-0.21)\\ 
\textbf{MUSCO} $\left(\text{nx}, 10, 3 \times 1/3\right)$ &  $1.77\times$  & 69.20 (-5.80) \\ 
\hline
\end{tabular*}
~
\caption{Compressed  Faster R-CNN (with ResNet-50 backbone) models on Pascal VOC2007 evaluation dataset. CPD-3 based compression is used. \textbf{MUSCO}$\left(\text{nx}, k, 3 \times 1/3\right)$ is a compressed model obtained after 3 compression steps (where 1/3 of layers is compressed per one step) using $k\times$ parameter reduction  at each step.}
\label{table:resnetvoc}
\end{table}

\subsection{Compression using SVD}
To compress a fully connected layer, we approximate its weight tensor $\theta\in\mathbb{R}^{l_\mathrm{in}\times l_\mathrm{out}}$ using rank-$R$ singular value decomposition (SVD). Namely, an approximation $\hat{\theta}$ can be represented as $\hat{\theta} = USV^T$, where $U\in\mathbb{R}^{l_\mathrm{in}\times R}$ and  $V\in\mathbb{R}^{l_\mathrm{out}\times R}$ are orthogonal matrices and $S$ is a diagonal matrix.

Thus, defining $\theta_{\mathrm{in}} = US$ and $\theta_{\mathrm{out}} = V^T$ we obtain
\begin{equation}
         \mathcal{F}_{\mathrm{fact}}(\hat{\theta}) = (\theta_{\mathrm{out}}, \theta_{\mathrm{in}}),
\end{equation}
and hence, the fully connected layer is replaced by two consecutive fully connected layers with weights $\theta_{\mathrm{in}}\in\mathbb{R}^{l_\mathrm{in}\times R}$ and $\theta_{\mathrm{out}}\in\mathbb{R}^{R \times l_\mathrm{out}}$. 
Automatic rank selection for rank-$R$ SVD is performed using GAS of EVBMF.

All models provided in Section~6.2 were iteratively compressed using Tucker-2 based compression for convolutional layers and SVD based for fully connected ones.



\subsection{Layer-by-layer statistics}
In Tables \ref{flops1} and \ref{flops2} there are shown floating points operation, required by each part of the initial and compressed models for both VGG and ResNet based models. 

\vspace{28pt}
\begin{table}[!h]
\onecolumn
\centering
\begin{tabular}{|c|c|l|l|l|l|l|}
\hline
\multirow{2}{*}{Layer} & \multirow{2}{*}{\begin{tabular}[c]{@{}c@{}}Res\\ Block\end{tabular}} & \multicolumn{2}{c|}{Original} & \multicolumn{3}{c|}{MUSCO (nx, 1.4, 3)} \\ \cline{3-7} 
 &  & \multicolumn{1}{c|}{\begin{tabular}[c]{@{}c@{}}In/out \\ channels\\ of (3x3)\\ kernel\end{tabular}} & \multicolumn{1}{c|}{\small{MFLOPs*}} & \multicolumn{1}{c|}{\begin{tabular}[c]{@{}c@{}}In/out \\ channels\\ of (3x3)\\ kernel\end{tabular}} & \multicolumn{1}{c|}{\small{MFLOPs*}} & \multicolumn{1}{c|}{\begin{tabular}[c]{@{}c@{}}\small{MFLOPs}\\ $\times$\end{tabular}} \\ \hline
\multicolumn{2}{|c|}{stem} & 3x64 (7x7) & 118 & 3x64 (7x7) & 118 & 1.00 \\ \hline
 & 0 & 64x64 & 231 & 25x25 & 143 & 1.61 \\
layer1 & 1 & 64x64 & 218 & 25x25 & 130 & 1.67 \\
 & 2 & 64x64 & 218 & 25x25 & 130 & 1.67 \\ \hline
\multirow{4}{*}{layer2} & 0 & 128x128 & 295 & 51x51 & 208 & 1.42 \\
 & 1 & 128x128 & 218 & 51x51 & 131 & 1.66 \\
 & 2 & 128x128 & 218 & 51x51 & 131 & 1.66 \\
 & 3 & 128x128 & 218 & 51x51 & 131 & 1.66 \\ \hline
 \multirow{6}{*}{layer3}& 0 & 256x256 & 295 & 103x103 & 209 & 1.41 \\
 & 1 & 256x256 & 218 & 103x103 & 132 & 1.66 \\
 & 2 & 256x256 & 218 & 103x103 & 132 & 1.66 \\
 & 3 & 256x256 & 218 & 103x103 & 132 & 1.66 \\
 & 4 & 256x256 & 218 & 103x103 & 132 & 1.66 \\
 & 5 & 256x256 & 218 & 103x103 & 132 & 1.66 \\ \hline
\end{tabular}
\\~
\caption{Faster R-CNN C4 with ResNet backbone fine-tuned on VOC2007+2012. In the table we represent each residual block by its conv2 layer with a kernel of spacial size~$3\times3$.  MFLOPs* are computed for all layers in a residual block, given an  input image of size $244\times 244\times 3$.}
\label{flops2}
\end{table}

\begin{table*}[]
\centering
\begin{tabular}{|c|l|l|l|l|l|}
\hline
\multicolumn{1}{|c|}{\multirow{2}{*}{Layer}} & \multicolumn{2}{c|}{Original shape} & \multicolumn{3}{c|}{MUSCO (nx, 1.77, 4)} \\ \cline{2-6} 
\multicolumn{1}{|c|}{} & \multicolumn{1}{c|}{\begin{tabular}[c]{@{}c@{}} In/out \\ channels\\ of (3x3)\\ kernel\end{tabular}} & \multicolumn{1}{c|}{\small{MFLOPs}} & \multicolumn{1}{c|}{\begin{tabular}[c]{@{}c@{}}In/out \\ channels,\\ (1x1)\\ (3x3)\\ (1x1)\\ kernels\end{tabular}} & \multicolumn{1}{c|}{\small{MFLOPs}} & \multicolumn{1}{c|}{\begin{tabular}[c]{@{}c@{}}\small{MFLOPs}\\ $\times$\end{tabular}} \\ \hline
1 & 3x64 & 87 & 3x64 & 87 & 1.00 \\ \hline
 &  &  & 64x16 & 32 &  \\
2 & 64x64 & 1850 & 16x16 & 45 & 16.91 \\
 &  &  & 16x64 & 32 &  \\ \hline
 &  &  & 64x17 & 8 &  \\
3 & 64x128 & 925 & 17x27 & 18 & 17.86 \\
 &  &  & 27x128 & 26 &  \\ \hline
 &  &  & 128x34 & 35 &  \\
4 & 128x128 & 1850 & 34x34 & 55 & 14.76 \\
 &  &  & 34x128 & 35 &  \\ \hline
 &  &  & 128x36 & 8 &  \\
5 & 128x256 & 925 & 36x57 & 21 & 15.76 \\
 &  &  & 57x256 & 28 &  \\ \hline
 &  &  & 256x69 & 35 &  \\
6 & 256x256 & 1850 & 69x69 & 55 & 14.76 \\
 &  &  & 69x256 & 35 &  \\ \hline
 &  &  & 256x69 & 35 &  \\
7 & 256x256 & 1850 & 69x69 & 55 & 14.76 \\
 &  &  & 69x256 & 35 &  \\ \hline
 &  &  & 256x73 & 9 &  \\
8 & 256x512 & 925 & 73x116 & 24 & 14.86 \\
 &  &  & 116x512 & 29 &  \\ \hline
 &  &  & 512x139 & 36 &  \\
9 & 512x512 & 1850 & 139x139 & 57 & 14.29 \\
 &  &  & 139x512 & 36 &  \\ \hline
 &  &  & 512x139 & 36 &  \\
10 & 512x512 & 1850 & 139x139 & 57 & 14.29 \\
 &  &  & 139x512 & 36 &  \\ \hline
 &  &  & 512x139 & 9 &  \\
11 & 512x512 & 462 & 139x139 & 14 & 14.29 \\
 &  &  & 139x512 & 9 &  \\ \hline
 &  &  & 512x139 & 9 &  \\
12 & 512x512 & 462 & 139x139 & 14 & 14.29 \\
 &  &  & 139x512 & 9 &  \\ \hline
 &  &  & 512x139 & 9 &  \\
13 & 512x512 & 462 & 139x139 & 14 & 14.29 \\
 &  &  & 139x512 & 9 &  \\ \hline
\end{tabular}
\\~
\caption{Faster R-CNN with VGG-16 backbone fine-tuned on VOC2007. In the table MFLOPs are computed for an input image of size $244\times244\times3$. MFLOPs~$\times$ denotes flops reduction for each  convolutional layer from the original model. MFLOPs~$\times$ is  calculated as  MFLOPs of original layer  devided by MFLOPs of   decomposed layer (i.e. a sum of MFLOPs of all three convolutional layers, substituting the original one). For example, for the original layer  with a weight of size $512\times512\times3\times3$, a decomposed layer consists of three layers with weights $139\times512\times1\times1$, $139\times139\times3\times3$, $512\times139\times1\times1$, respectively.}
\label{flops1}
\end{table*}

\end{document}